\documentclass[conference]{IEEEtran}
\IEEEoverridecommandlockouts

\usepackage{amsmath,amssymb,amsfonts}
\usepackage{algorithmic}
\usepackage{graphicx}
\usepackage{textcomp}
\usepackage{xcolor}
\usepackage{multirow}
\usepackage{siunitx}
\usepackage{booktabs}
\def\BibTeX{{\rm B\kern-.05em{\sc i\kern-.025em b}\kern-.08em
    T\kern-.1667em\lower.7ex\hbox{E}\kern-.125emX}}
\usepackage[colorlinks=true,linkcolor=blue,citecolor=blue]{hyperref}
\usepackage[style=numeric,sorting=none]{biblatex}
\begin{document}

\title{Leveraging Transformer Decoder for Automotive Radar Object Detection}

\author{
\IEEEauthorblockN{
Changxu Zhang\textsuperscript{1, 3},
Zhaoze Wang\textsuperscript{1, 3},
Tai Fei\textsuperscript{2},
Christopher Grimm\textsuperscript{1},
Yi Jin\textsuperscript{1},
Claas Tebruegge\textsuperscript{1}, \\
Ernst Warsitz\textsuperscript{1},
Markus Gardill\textsuperscript{3}
}
\IEEEauthorblockA{
\textsuperscript{1}\textit{HELLA GmbH \& Co. KGaA}, Lippstadt, Germany \\
\textsuperscript{2}\textit{Dortmund University of Applied Sciences and Arts}, Dortmund, Germany \\
\textsuperscript{3}\textit{Brandenburg University of Technology}, Cottbus, Germany \\
Email: \{changxu.zhang, zhaoze.wang, christopher.grimm, yi.jin, claas.tebruegge, ernst.warsitz\}@forvia.com, \\
tai.fei@fh-dortmund.de, markus.gardill@b-tu.de}
}
\maketitle

\begin{abstract}

In this paper, we present a Transformer-based architecture for 3D radar object detection that uses a novel Transformer Decoder as the prediction head to directly regress 3D bounding boxes and class scores from radar feature representations. To bridge multi-scale radar features and the decoder, we propose Pyramid Token Fusion (PTF), a lightweight module that converts a feature pyramid into a unified, scale-aware token sequence. By formulating detection as a set prediction problem with learnable object queries and positional encodings, our design models long-range spatial–temporal correlations and cross-feature interactions. This approach eliminates dense proposal generation and heuristic post-processing such as extensive non-maximum suppression (NMS) tuning. We evaluate the proposed framework on the RADDet, where it achieves significant improvements over state-of-the-art radar-only baselines.

\end{abstract}


\section{Introduction}
\label{sec:intro}

Millimeter-wave radar is a key sensing modality in modern perception systems for advanced driver-assistance systems (ADAS) and autonomous vehicles. Compared with optical sensors, the millimeter-wave radar, which directly measures object range and relative radial velocity, provides robust performance under adverse weather and low-visibility conditions because its electromagnetic waves penetrate fog, rain, and airborne particulates. These properties make radar an indispensable complement to cameras and LiDAR in safety-critical tasks such as Adaptive Cruise Control (ACC) and Automatic Emergency Braking (AEB).

Accurate and reliable object detection, including identifying and localizing vehicles, pedestrians, cyclists, and static obstacles, is fundamental to radar-based perception. However, processing radar data presents multiple challenges, as the data are sparse, noisy, and severely affected by artifacts caused by multipath reflection. These characteristics, coupled with the diversity of radar data representations, demand dedicated architectures that can effectively capture spatial and temporal dependencies across multiple dimensions.

Transformer-based architectures have recently achieved remarkable success in vision~\cite{carion2020detr} \cite{zhu2020deformabledetr}, 
yet their potential for radar perception remains relatively underexplored. Recently, several radar-detection models relying on convolutional neural networks (CNNs) or classical signal-processing pipelines have been proposed. They excel at modeling local patterns, but struggle to capture long-range dependencies and global spatial relations. In contrast, the attention mechanism~\cite{vaswani2017attention} of transformers is inherently well-suited for modeling global context across range, Doppler, and angle dimensions, enabling more robust reasoning in cluttered or ambiguous radar scenes.

In this work, we present a Transformer-based architecture for 3D radar object detection that employs a novel Transformer Decoder head for end-to-end prediction. The proposed framework directly regresses 3D bounding boxes and class scores from radar feature representations, completely getting rid of reliance on hand-crafted proposals and heuristic post-processing.
Specifically, we incorporate a Transformer Decoder that takes advantage of learnable object queries and positional encodings to jointly model long-range dependencies and cross-feature interactions. To efficiently integrate multi-scale spatial information, we propose Pyramid Token Fusion (PTF), a lightweight module that converts hierarchical feature maps into a unified, scale-aware token sequence for the decoder. This design avoids the need for dense proposal generation and non-maximum suppression (NMS), resulting in a cleaner and more systematic end-to-end optimization pipeline. Comprehensive experiments on the RADDet dataset~\cite{zhang2021raddet} demonstrate that our approach achieves significant improvements over existing radar-based object detection methods.
The remainder of this paper is organized as follows: Section~\ref{sec:relatedwork} reviews related detection methods, Section~\ref{sec:methodology} presents the proposed model, Section~\ref{sec:experiments} outlines the experiments, and Section~\ref{sec:conclusion} concludes with limitations and future work.

\section{Related Works}
\label{sec:relatedwork}
This section reviews previous studies on vision- and radar-based object detection, organized into three categories: CNN-based radar detectors, DETR-based vision detectors, and recent transformer-based radar detection frameworks.

\subsection{CNN-based Radar Object Detection}

Early radar-based object detection approaches primarily relied on convolutional architectures inspired by vision models. Major et al.~\cite{major2019} applied 3D CNNs to Range-Azimuth-Doppler (RAD) tensors, introducing a polar-to-Cartesian transformation to align radar features with the Bird's-Eye View (BEV) detection space. Zhang et al.~\cite{zhang2021raddet} proposed RADDet, adapting a ResNet-style~\cite{he2016resnet} backbone and a YOLO-based detection head~\cite{bochkovskiy2020yolov4} for 3D radar detection, establishing a widely adopted baseline. Gao et al.~\cite{gao2020rampcnn} introduced RAMP-CNN, which decomposes the radar cube into Range–Angle (RA), Range-Doppler (RD) and Angle–Doppler (AD) projections, each processed by 3D convolutional autoencoders before feature fusion. Wang et al.~\cite{wang2021rodnet} further proposed RODNet, trained under a camera–radar cross-modal supervision scheme. 
Jin et al. \cite{jin2023cross} proposed a 2D CNN-based network that employs the complex RD matrix as input to achieve radar-tailored panoptic segmentation, i.e., free-space segmentation and object detection, in the camera image.

\subsection{DETR-based Object Detection}

Transformers revolutionized object detection by reformulating it as a set prediction task. The original DETR~\cite{carion2020detr} employs a transformer encoder–decoder to directly predict object sets using learned queries and bipartite matching, eliminating anchors and NMS. However, it suffered from slow convergence and poor multi-scale handling. Deformable DETR~\cite{zhu2020deformabledetr} addressed these issues by introducing sparse multi-scale attention around reference points, greatly improving convergence and scalability. Meanwhile, hierarchical backbones such as Swin Transformer~\cite{liu2021swin} provided efficient multi-scale feature extraction through shifted-window self-attention, enhancing DETR’s capability on small objects. Conditional DETR~\cite{meng2021conditionaldetr} further stabilized training by conditioning attention on learned reference points and separating content and positional embeddings, yielding sharper localization and faster optimization.

\subsection{Transformer-based Radar Object Detection}

Building on RODNet, transformer-based radar detectors leverage attention mechanisms to capture long-range dependencies and temporal context. T-RODNet~\cite{jiang2022trodnet} and SS-RODNet~\cite{zhuang2023ssrodnet} integrate 3D Swin Transformer blocks to model hierarchical spatio-temporal features directly from RAD tensors, improving detection under occlusion and clutter. 
Besides, TransRAD~\cite{cheng2025transrad} introduces a retentive Manhattan self-attention (MaSA) mechanism to enhance spatial saliency, combining transformer feature extraction with a CNN prediction head and location-aware NMS for refined localization. 
However, such CNN-based heads still rely on hand-crafted post-processing, which limits end-to-end optimization. To overcome this, our framework adopts a transformer decoder–based prediction head inspired by DETR, enabling direct query-based object prediction and fully end-to-end radar detection.

\section{Methodology}
\label{sec:methodology}
\begin{figure*}[htb]
  \centering
  \centerline{\includegraphics[width=18cm]{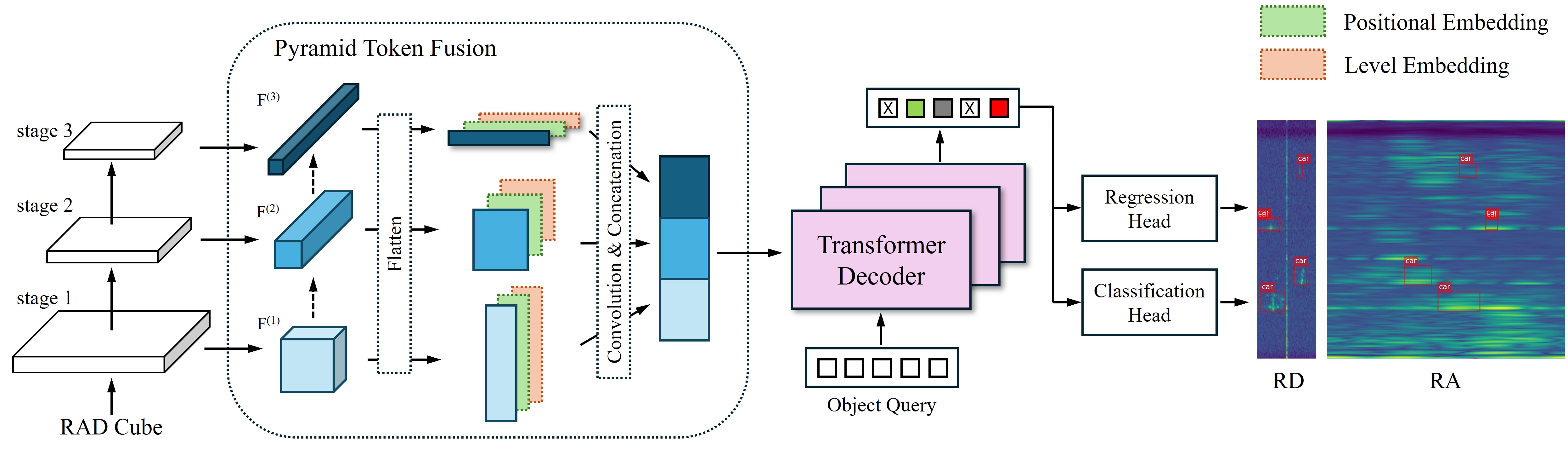}} 
\caption{Overall architecture of the proposed model. The RAD cube is encoded by a hierarchical transformer backbone to produce multi-scale features, which are augmented with positional and level embeddings in the PTF module and fused into a unified token sequence. This sequence serves as memory for the transformer decoder, whose object queries generate class scores and 3D bounding box predictions via two lightweight heads.}
\label{figure:1}
\end{figure*}
As illustrated in Fig. \ref{figure:1}, the proposed framework adopts a transformer-based backbone from TransRAD \cite{cheng2025transrad} to extract hierarchical features from the input 3D RAD cube. Multi-scale features are unified through a PTF module, which aligns channel dimensions and encodes spatial and scale information. The fused tokens are processed by a conditional transformer decoder driven by learnable object queries. Finally, two lightweight feed-forward networks (FFNs) predict class probabilities and 3D bounding boxes, forming an end-to-end, anchor-free, and NMS-free detection architecture.

\subsection{Pyramid Token Fusion}
Transformer decoders benefit from attending to rich multi-scale context. However, directly fusing feature pyramids often leads to mismatched channel dimensions, unbalanced token counts, and scale ambiguity. To address these issues, we propose PTF, a lightweight and parameter-efficient module that unifies a feature pyramid into a single, scale-aware token sequence for the decoder. Specifically, PTF performs three key operations: it aligns channels across different levels, injects both spatial and level positional information, and flattens and concatenates all levels into one contiguous memory for cross‑attention.

Let ${\mathbf{F}^{(l)}\in \mathbb{R}^{C_l \times H_l \times W_l}}$ denote feature maps extracted from different pyramid levels, where $C_l$ is the number of channels and $(H_l, W_l)$ represent the height and width of feature map at level $l$. The total number of spatial positions is $N_l = H_l W_l$ for each level and $N = \sum_l N_l$ across all levels. To establish a uniform representation space, each feature map is projected to a shared embedding dimension $D$ through a $1{\times}1$ convolution followed by layer normalization, yielding $\tilde{\mathbf{F}}^{(l)}\in \mathbb{R}^{D \times H_l \times W_l}$. This step harmonizes channel dimensions while preserving the spatial layout of each level.


Beyond channel alignment, preserving positional and scale information is essential for effective multi-scale fusion. We therefore augment $\tilde{\mathbf{F}}^{(l)}$ with two complementary encodings: a spatial positional encoding $\mathbf{PE}^{(l)} \in \mathbb{R}^{D \times H_l \times W_l}$ that captures intra-level location cues, and a learnable level embedding $\mathbf{e}^{(l)} \in \mathbb{R}^{D \times H_l \times W_l}$ that represents the relative scale of each pyramid level and is broadcast across its spatial grid:
\begin{equation}
    \hat{\mathbf{F}}^{(l)} = \tilde{\mathbf{F}}^{(l)} + \mathbf{PE}^{(l)} + \mathbf{e}^{(l)}.
\end{equation}

To further enhance spatial discrimination along radar-specific dimensions, we adopt a tunable positional encoding (TPE) scheme inspired by \cite{yataka2024retr}, allowing adaptive emphasis between Doppler and azimuth axes. Let the total positional embedding dimension $d_{\text{pos}}$ be a subset of the feature dimension $D$ (i.e., $d_{\text{pos}} \leq D$). We decompose it as:
\begin{equation}
\begin{aligned}
    d_{\text{dop}} &= \alpha d_{\text{pos}}, \quad
    d_{\text{azi}} = (1 - \alpha)d_{\text{pos}}, \\
    &\Rightarrow\; d_{\text{dop}} + d_{\text{azi}} = d_{\text{pos}},
\end{aligned}
\end{equation}
where $\alpha \in [0, 1]$ controls the relative weighting of Doppler versus azimuth information.

Finally, each encoded feature map $\hat{\mathbf{F}}^{(l)}$ is rasterized into a sequence of $N_l$ tokens through column-major flattening, and all levels are concatenated to form the final unified token memory $\mathbf{F}^{\mathrm{all}}\in \mathbb{R}^{N \times D}$. This fused representation preserves spatial locality through $\mathbf{PE}^{(l)}$ and maintains scale awareness via $\mathbf{e}^{(l)}$, providing the decoder with a compact yet information-rich representation of the radar scene.

\subsection{Transformer Decoder Head}
Conventional CNN-based prediction heads, such as those used in YOLO-style detectors, generate dense, grid-based predictions and rely on heuristic post-processing steps like NMS and anchor or stride tuning to obtain the final detections. These heuristics introduce additional hyperparameters and often struggle in cluttered radar scenes where reflections overlap or objects are closely spaced. Furthermore, CNN-based heads inherently rely on local receptive fields, which constrain their ability to capture long-range dependencies and global spatial relationships. This limitation is particularly problematic for radar perception, where multipath propagation and ghost targets are common, requiring broader contextual reasoning to distinguish genuine object responses from spurious reflections.

In contrast, transformer-based decoders formulate detection as an end-to-end set prediction problem \cite{carion2020detr}. By maintaining a fixed number of object queries and employing one-to-one bipartite matching during training, such models naturally avoid the need for NMS. The attention mechanism \cite{vaswani2017attention} allows for global context modeling across the scene, enhancing robustness under occlusion and multipath distortion. Furthermore, the decoder’s iterative refinement of object queries progressively improves localization accuracy across layers.

Conditional DETR~\cite{meng2021conditionaldetr} enhances the original DETR design with faster convergence and improved spatial precision. Each decoder query is decomposed into a content component (what to detect) and a positional component (where to attend), where cross-attention is conditioned on a learnable reference point. The positional component is derived from both the reference point $\mathbf{s}$ and the query feature $\mathbf{f}$, mapped jointly into a positional query embedding:
\begin{equation}
    (\mathbf{s}, \mathbf{f}) \rightarrow \mathbf{p}_q.
\end{equation}
This formulation constrains the cross-attention to a local region around the reference point while preserving semantic flexibility, stabilizing training and improving localization accuracy. 

Each decoder layer follows the standard transformer structure composed of self-attention, cross-attention, and a FFN. Let $\mathbf{Q}^k$ denote the set of object queries at layer $k$.
The queries first exchange contextual information through self-attention: 
\begin{align}
    \tilde{\mathbf{Q}}^k &= \mathbf{Q}^k
      + \mathtt{Att}_{\mathtt{self}}\!\left(
          \mathtt{Que}(\mathbf{Q}^k),\,
          \mathtt{Key}(\mathbf{Q}^k),\,
          \mathtt{Val}(\mathbf{Q}^k)
        \right),
\end{align}

\noindent then attend to the encoded radar features via cross-attention
\begin{align}
    \hat{\mathbf{Q}}^k &= \tilde{\mathbf{Q}}^k
          + \mathtt{Att}_{\mathtt{crs}}\!\left(
              \mathtt{Que}(\tilde{\mathbf{Q}}^k),\,
              \mathtt{Key}\!\left(\mathbf{F}^{\mathrm{all}}\right),\,
              \mathtt{Val}\!\left(\mathbf{F}^{\mathrm{all}}\right)
            \right),
\end{align}
and are finally refined by the FFN:
\begin{align}
    \mathbf{Q}^{k+1} &= \hat{\mathbf{Q}}^k
      + \mathrm{FFN}(\hat{\mathbf{Q}}^k).
\end{align} 

The regression and classification heads are lightweight FFNs applied to the decoder output. The regression head predicts 3D bounding box parameters (position, size, and Doppler) relative to the reference point, followed by a sigmoid activation to constrain predictions within valid spatial ranges.
The classification branch maps each query embedding to class logits, producing probabilities over all object classes, including the “no object” category. 
The predicted 3D boxes are further decomposed into RA and RD, enabling auxiliary supervision across multiple radar views.

\subsection{Loss Function}
Over the permutation set $\mathcal{S}_M$ of $M$ predictions and ground-truth objects, the Hungarian algorithm \cite{kuhn1955hungarian} is used with a matching cost matrix to determine the optimal assignment $\sigma^{*} \in \mathcal{S}_M$. The loss is then computed exclusively on these matched pairs, referred to as the set-prediction loss. 

For bounding box regression, we combine the Generalized Intersection over Union (GIoU) loss \cite{Rezatofighi2018giou} $\mathcal{L}_{GIoU}$ and the $\ell_1$ loss $\mathcal{L}_{L1}$. The GIoU loss extends standard IoU by considering the smallest enclosing box of the prediction and ground truth, providing meaningful gradients even when boxes do not overlap:
\begin{equation}
\mathcal{L}_{GIoU} = 1 - \text{IoU} + \frac{|C \setminus (B_p \cup B_{gt})|}{|C|},
\end{equation}

\noindent where $B_p$ and $B_{gt}$ are the predicted and ground-truth bounding boxes, $C$ is the smallest enclosing box covering both, and $|\cdot|$ denotes the box area. The coefficients $\beta_{GIoU}$ and $\beta_{L1}$ are weighting factors that balance the contributions of the GIoU and $\ell_1$ losses. The combined bounding box loss is then:
\begin{equation}
\label{eq:bboxloss}
\mathcal{L}_{Bbox} = \beta_{GIoU}\mathcal{L}_{GIoU} + \beta_{L1}\mathcal{L}_{L1},
\end{equation}

For classification, we use Focal Loss \cite{lin2017focalloss} $\mathcal{L}_{Class}$, which reduces the influence of easy examples and focuses training on hard, misclassified instances:
\begin{equation}
\mathcal{L}_{Class} = -\alpha_t (1 - p_t)^\gamma \log(p_t),
\end{equation}

\noindent where $p_t$ is the predicted probability for the true class, $\alpha_t$ balances class weights, and $\gamma$ is the focusing parameter controlling the down-weighting of easy examples.


Finally, the total loss is defined as a weighted sum of the following components:
\begin{equation}
\label{eq:totalloss}
\mathcal{L} = \beta_1\mathcal{L}_{Bbox}^{RAD} + \beta_2\mathcal{L}_{Bbox}^{RA} + \beta_3\mathcal{L}_{Bbox}^{RD} + \beta_4 \mathcal{L}_{Class},
\end{equation}
where $\beta_1$, $\beta_2$, $\beta_3$, and $\beta_4$ are weighting coefficients that control the relative importance of each loss term in the overall optimization.

\section{Experiments}
\label{sec:experiments}

\subsection{Dataset}
In this work, we use RADDet \cite{zhang2021raddet} to train and evaluate our method. RADDet provides synchronized 3D radar cubes 
collected using a Texas Instruments AWR1843-BOOST radar 
in urban road environments under normal weather conditions, without ego-motion. The dataset comprises 10,158 radar frames with 28,401 annotated objects across six classes (person, bicycle, car, motorcycle, bus, and truck), where each instance is labeled with a 3D bounding box defined by its center coordinates and dimensions. The annotation utilizes a combination of depth estimation from stereo images and Constant False Alarm Rate (CFAR) detection on the radar spectrum, enabling accurate 3D localization including Doppler information. Each radar cube has a shape of $(256, 256, 64)$ and contains complex-valued measurements. 

\subsection{Evaluation Metrics}
To quantitatively evaluate the proposed radar object detection model, we adopt the mean Average Precision (mAP) metric, which summarizes both detection accuracy and localization quality. For each object class $c$, the Average Precision (AP) is defined as the area under the precision-recall curve, estimated by a discrete summation over recall intervals \cite{padilla2020metrics}:
\begin{equation}
\text{AP} = \int_{0}^{1} P(R) \, dR \approx \sum_{n} (R_{n+1} - R_n) \max(P_{n+1}, P_n),
\end{equation}

\noindent where $P_n$ and $R_n$ are the precision and recall at the $n^{\text{th}}$ threshold. The mAP is then computed as the average over all $C$ classes:
\begin{equation}
\text{mAP} = \frac{1}{C} \sum_{c=1}^{C} \frac{1}{N} \sum_{n=1}^{N} \text{AP}_{c}(\text{IoU}_{n}),
\end{equation}



\noindent where $\text{AP}_{c}(\text{IoU}_{n})$ denotes the AP for the $c^{\text{th}}$ class evaluated at the $n^{\text{th}}$ IoU threshold, $N$ is the total number of IoU thresholds.
For our evaluation, mAP is computed separately for the RA, RD, and full 3D RAD box.

\subsection{Training Details}
All experiments were conducted on a single NVIDIA RTX 4090 GPU with an Intel i9-14900KF CPU and 64 GB of RAM. The dataset was split into 80\% for training and 20\% for testing, and no data augmentation techniques were applied. For preprocessing, the magnitude of the complex-valued radar tensors was computed, followed by a $\log_{10}$ transformation to compress the dynamic range and enhance feature visibility. The model was trained for 150 epochs with a batch size 8 using the AdamW optimizer with an initial learning rate of $1\times10^{-4}$ and standard weight decay. Training is performed end-to-end without any post-processing. 

Our backbone is adopted from the of TransRAD \cite{cheng2025transrad}, which introduces a transformer-based encoder designed to capture irregular object shapes and spatial saliency in radar data. The network consists of four encoder layers and three decoder layers, and the feature pyramid includes maps at resolutions of 32, 16, and 8. For PTF, we set $D=128$ for channel alignment and $\alpha=0.6$ for TPE. The Hungarian matcher assigns predicted queries to ground-truth objects based on a combined cost of classification and bounding box regression. The total loss weights are set as $[\beta_1, \beta_2, \beta_3, \beta_4, \beta_{GIoU}, \beta_{L1}] = [40, 15, 15, 10, 5, 5]$.

\subsection{Benchmark Comparison and Performance Evaluation} 
Table~\ref{tab:3dresult} and~\ref{tab:2dresult} summarize the performance of our proposed method against several benchmark radar-based 3D and 2D detection approaches, including both baseline and state-of-the-art (SOTA) models. RADDet \cite{zhang2021raddet} serves as our baseline. For 3D detection, evaluation is conducted using mAP at multiple IoU thresholds ($0.4$–$0.7$) for RAD 3D bounding boxes, while 2D detection adopts stricter IoU thresholds ($0.5$–$0.9$) to better assess localization precision. All methods are categorized into CNN-based and Transformer-based approaches, separated by a horizontal line, with the best results highlighted in bold. To ensure fair comparison, the performance of reference methods is reported as presented in TransRAD~\cite{cheng2025transrad}.

\begin{table}[t]
\centering
\caption{Comparison of different models for RAD 3D Detection}
\label{tab:3dresult}
\begin{tabular}{l c r r r r}
\toprule
\multirow{2}{*}{Method} & Params $\downarrow$ & \multicolumn{4}{c}{mAP(\%) $\uparrow$ for RAD 3D Detection} \\
\cmidrule(lr){2-2} \cmidrule(lr){3-6}
 & (M) & $AP_{0.4}$ & $AP_{0.5}$ & $AP_{0.6}$ & $AP_{0.7}$ \\
\midrule
RADDet \cite{zhang2021raddet} & 8.07 & 34.83 & 19.41 & 8.66 & 2.78 \\
RODNet \cite{wang2021rodnet} & 36.89 & 37.74 & 24.53 & 12.71 & 5.30 \\
RAMP-CNN \cite{gao2020rampcnn} & 107 & 31.63 & 19.29 & 9.39 & 4.16 \\
\midrule
T-RODNet \cite{jiang2022trodnet} & 161 & 15.91 & 10.81 & 6.62 & 3.37 \\
RadarFormer \cite{dalbah2023radarformer} & 6.19 & 22.80 & 14.33 & 8.05 & 3.83 \\
TransRAD \cite{cheng2025transrad} & \textbf{5.78} & 50.89 & 38.76 & 25.54 & 13.17 \\
\midrule
\textbf{Ours} & 6.37 & \textbf{53.75} & \textbf{41.38} & \textbf{28.56} & \textbf{15.91} \\
\bottomrule
\end{tabular}
\end{table}

\begin{table*}[t]
\centering
\caption{Comparison of different models for RA and RD 2D Detection} 
\label{tab:2dresult}
\begin{tabular}{lrrrrrrrrrr}
\toprule
\multirow{2}{*}{Method} & \multicolumn{5}{c}{mAP(\%) $\uparrow$ for RA 2D Detection} & \multicolumn{5}{c}{mAP(\%) $\uparrow$ for RD 2D Detection} \\
\cmidrule(lr){2-6} \cmidrule(lr){7-11}
& $AP_{0.5}$ & $AP_{0.6}$ & $AP_{0.7}$ & $AP_{0.8}$ & $AP_{0.9}$ & $AP_{0.5}$ & $AP_{0.6}$ & $AP_{0.7}$ & $AP_{0.8}$ & $AP_{0.9}$ \\
\midrule
RADDet \cite{zhang2021raddet} & 45.35 & 29.29 & 14.78 & 4.05 & 0.29 & 39.45 & 24.73 & 13.33 & 4.19 & 0.49 \\
RODNet \cite{wang2021rodnet}  & 46.38 & 34.52 & 19.88 & 6.37 & 0.63 & 39.23 & 26.61 & 15.97 & 6.78 & 1.59 \\
RAMP-CNN \cite{gao2020rampcnn}     & 39.72 & 27.33 & 14.56 & 4.48 & 0.38 & 33.60 & 20.86 & 10.95 & 4.75 & 0.75 \\
\midrule
T-RODNet \cite{jiang2022trodnet}    & 18.98 & 15.36 & 11.69 & 3.43 & 0.13 & 21.98 & 13.90 & 6.25 & 2.75 & 0.75 \\
RadarFormer \cite{dalbah2023radarformer} & 31.38 & 23.82 & 16.25 & 4.18 & 0.20 & 28.30 & 13.52 & 7.86 & 3.49 & 0.96 \\
TransRAD \cite{cheng2025transrad} & \textbf{55.90} & \textbf{45.78} & \textbf{32.16} & 14.27 & 1.64 & 51.80 & 40.27 & 27.37 & 14.16 & 3.63 \\
\midrule
\textbf{Ours} & 55.38 & 44.87 & 30.62 & \textbf{15.04} & \textbf{2.03} & \textbf{55.91} & \textbf{44.89} & \textbf{31.72} & \textbf{16.25} & \textbf{4.76} \\
\bottomrule
\end{tabular}
\end{table*}

\begin{figure*}[ht]
  \centering
  \centerline{\includegraphics[width=18cm]{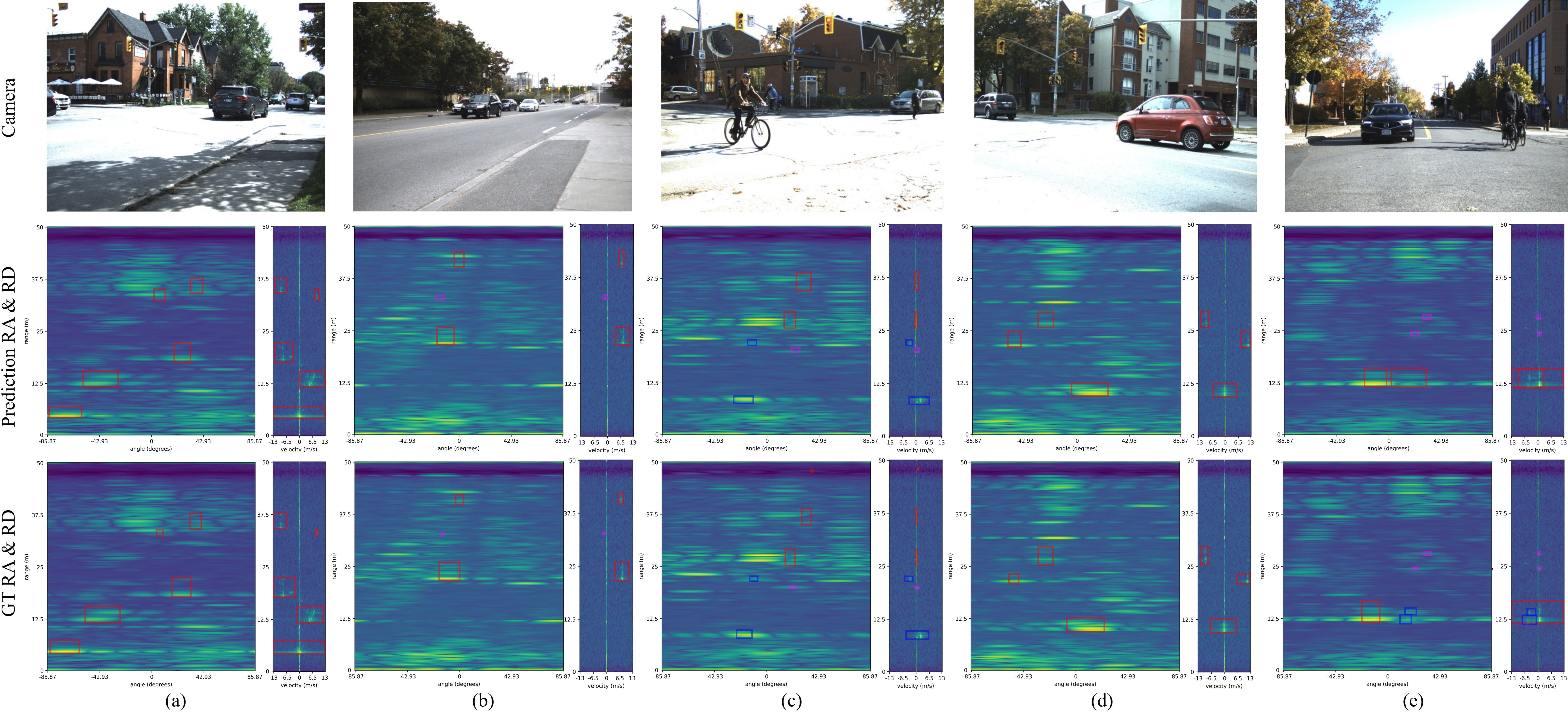}} 
\caption{Examples of detection results from our model shown for five different scenes (a-e). The top row displays the camera images. The middle row shows our model's predictions, overlaid on the RA map (left, x: azimuth, y: range) and RD map (right, x: Doppler, y: range). The bottom row presents the corresponding ground-truth annotations. Bounding box colors indicate object classes: red for cars, purple for pedestrians, and blue for cyclists.}
\label{figure:2}
\end{figure*}


As shown in Table~\ref{tab:3dresult}, the proposed model achieves the highest mAP across all IoU thresholds, outperforming all existing baselines on 3D radar object detection. Compared to the strong TransRAD~\cite{cheng2025transrad}, our approach improves mAP by $2.86\%$, $2.62\%$, $3.02\%$, and $2.74\%$ at IoU thresholds of 0.4, 0.5, 0.6, and 0.7, respectively, indicating more accurate localization and stronger generalization to challenging radar scenes. Despite having a comparable parameter count (6.37M vs. 5.78M), our model delivers superior precision, highlighting the efficiency of our PTF and decoder design. These results confirm that the proposed transformer-based framework effectively captures global spatio-temporal dependencies in radar data, leading to robust and precise 3D object detection.

As summarized in Table~\ref{tab:2dresult}. Our model achieves the best mAP at most IoU thresholds for both RA and RD 2D detections, outperforming existing methods at higher thresholds. For RA detection, while our mAP is slightly lower than TransRAD at the lower IoU thresholds ($0.5$–$0.7$) due to hyperparameter choices such as $\alpha$ in TPE, it surpasses all baselines at stricter thresholds ($0.8$–$0.9$), indicating more precise localization. On RD maps, our method consistently achieves the highest mAP across all thresholds, highlighting its ability to accurately capture Doppler information. These results demonstrate that the proposed architecture effectively leverages radar-specific features and global context modeling to improve both 2D and 3D detection performance.

\subsection{Qualitative Result Analysis}
Fig. \ref{figure:2} offers a qualitative visualization of our model's detection performance across five traffic scenarios (a-e). Each column compares the camera view (top) with our model's predictions on the RA \& RD maps (middle) and the corresponding ground truth (bottom). Visually, our model's predictions closely align with the ground truth across all examples, particularly for vulnerable road users, such as the occluded pedestrian in scene (b) and the cyclist in scene (c).
Furthermore, the model reliably detects vehicles at various ranges and in different maneuvers, as shown in scenes (a), (b), and (d). This visual evidence corroborates the strong quantitative mAP scores presented in the Table \ref{tab:3dresult} and \ref{tab:2dresult}, highlighting the model's precise localization capabilities. Some limitations are also observed: in (e), two bicycles are incorrectly merged into a single car detection, likely due to similar four-wheel reflection patterns; and the distant vehicle in (c) is missed, possibly because it lies near the radar’s maximum effective range. It is worth noting that due to the difference in the field of view (FoV) between the radar and camera, certain objects visible to radar may not appear in camera images, which can lead to apparent discrepancies.

\begin{figure}[h]
  \centering
  \centerline{\includegraphics[width=8cm]{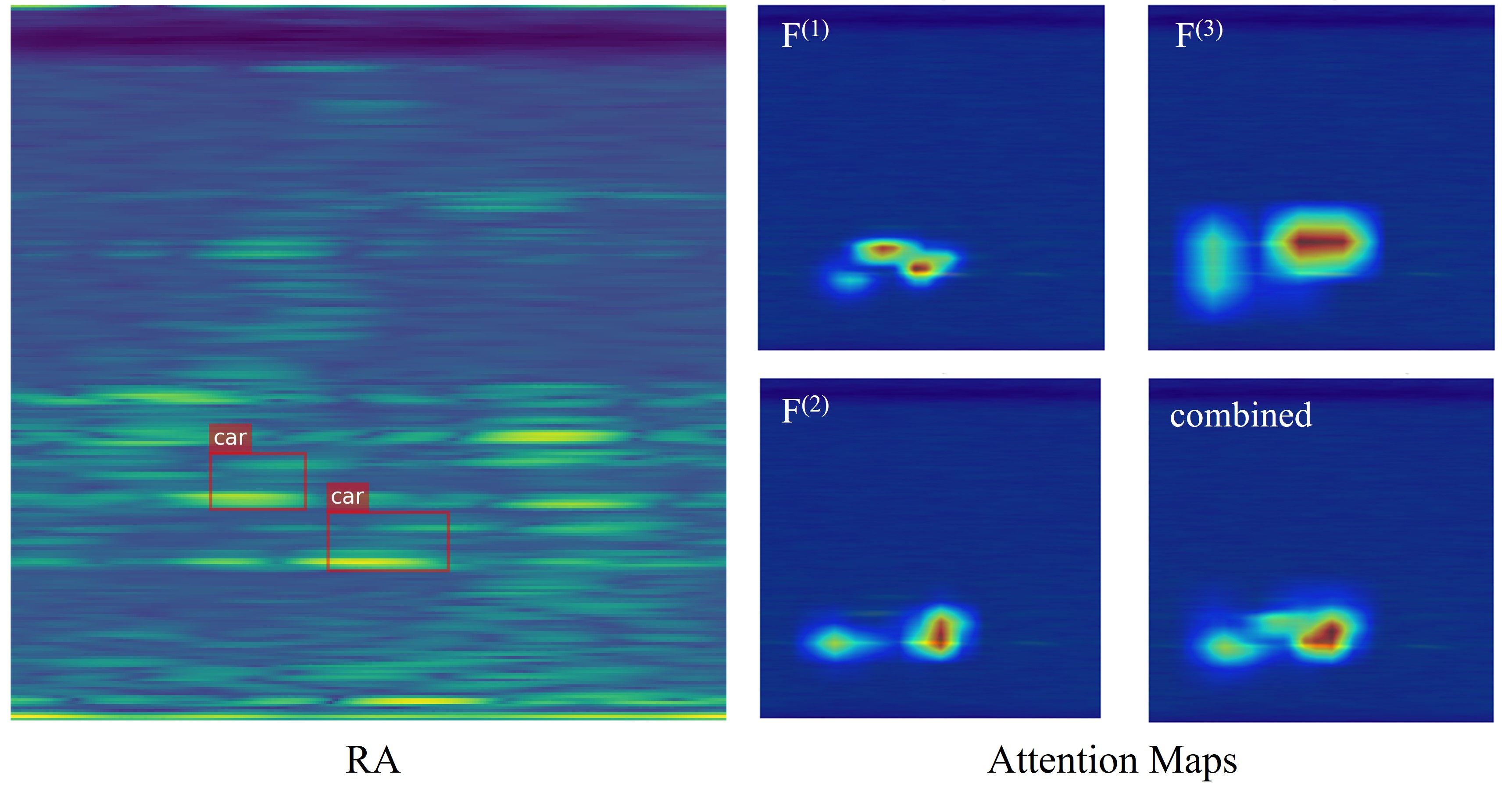}} 
\caption{Cross-attention maps on multi-scale feature maps (right), shown alongside RA (left) map with predicted bounding boxes. The spatial dimensions of the attention maps correspond to those of the RA map.}
\label{figure:3}
\end{figure}

Fig.~\ref{figure:3} visualizes the cross-attention map at the final decoder layer between object queries and the fused multi-scale feature maps. The attention patterns show that the prediction head consistently concentrates on the spatially relevant radar returns across scales, enabling precise localization of object centers and effective separation of nearby or overlapping echoes. These maps therefore provide qualitative evidence that the query-based decoder leverages multi-scale context to produce accurate, well-localized detections.

\section{Conclusion}
\label{sec:conclusion}
In this paper, we introduce a novel Transformer-based framework for 3D radar object detection, featuring a Transformer Decoder head for end-to-end set prediction and a PTF module for efficient multi-scale feature unification. This NMS-free architecture achieves SOTA performance on the RADDet dataset, delivering notably higher localization precision, particularly at stricter IoU thresholds. The main contribution of this work lies in validating the feasibility and effectiveness of Transformer Decoder architectures for automotive radar perception, aiming to inspire further transformer-based radar research. Nonetheless, our current evaluation is limited to a single dataset without ego-motion and elevation information, and performance may be constrained by annotation quality. Future work will focus on extending validation to more diverse datasets and improving label reliability.

\section*{Acknowledgment} 
This work is partially funded by the research project "NXT GEN AI Methods", supported by the German Federal Ministry for Economic Affairs and Energy (BMWE).


\end{document}